\pdfoutput=1

\documentclass[letterpaper,10pt,conference]{ieeeconf}

\IEEEoverridecommandlockouts
\usepackage{newtxtext}
\usepackage{etoolbox}
\makeatletter
\patchcmd{\@makecaption}{\scshape}{}{}{}
\makeatother
\usepackage{amsmath}
\usepackage{amssymb}
\usepackage{booktabs}
\usepackage{array}
\usepackage{cite}
\usepackage{graphicx}
\usepackage{eso-pic}
\usepackage{placeins}
\usepackage{xcolor}
\usepackage{colortbl}
\usepackage{url}

\usepackage[capitalize]{cleveref}
\crefname{section}{Sec.}{Secs.}
\Crefname{section}{Section}{Sections}
\Crefname{table}{Table}{Tables}
\crefname{table}{Tab.}{Tabs.}

\definecolor{TableAll}{gray}{0.88}
\definecolor{TableYes}{RGB}{0,128,82}
\definecolor{TablePartial}{RGB}{181,101,0}
\definecolor{TableNo}{RGB}{190,45,45}
\newcommand{\tblYes}{\textcolor{TableYes}{\ensuremath{\checkmark}}}
\newcommand{\tblPartial}{\textcolor{TablePartial}{\ensuremath{(\checkmark)}}}
\newcommand{\tblNo}{\textcolor{TableNo}{\ensuremath{\times}}}
\newcommand{\tblNR}{\tblNo}

\newcommand{\sysname}{BT-Agent}
\definecolor{RoleContext}{HTML}{0072B2}
\definecolor{RoleAction}{HTML}{E69F00}
\definecolor{RoleTerminal}{HTML}{009E73}
\definecolor{RoleHarness}{HTML}{CC79A7}
\definecolor{StateOk}{HTML}{009E73}
\definecolor{StateFail}{HTML}{D55E00}
\newcommand{\ctx}[1]{\textcolor{RoleContext}{\texttt{#1}}}
\newcommand{\act}[1]{\textcolor{RoleAction}{\texttt{#1}}}
\newcommand{\term}[1]{\textcolor{RoleTerminal}{\texttt{#1}}}

\title{\LARGE\bf
BT-Agent: Deploying Onboard Robot Agents with Behavior Trees and Multi-Fidelity Evaluation
}

\author{Fabio H{\"u}bel$^{1}$, Adam Lilja$^{1}$, Yue Yin$^{1}$, Siming He$^{2}$,\\
Claire Tomlin$^{2}$, Shankar Sastry$^{2}$, Jonas Frey$^{1,2}$, Marco Pavone$^{1,3}$%
\thanks{$^{1}$Stanford University. $^{2}$University of California, Berkeley.
$^{3}$NVIDIA.}%
}

\begin{document}

\IEEEaftertitletext{%
  \refstepcounter{figure}\label{fig:architecture}%
  \parbox{\textwidth}{%
    \centering
    \includegraphics[width=\textwidth]{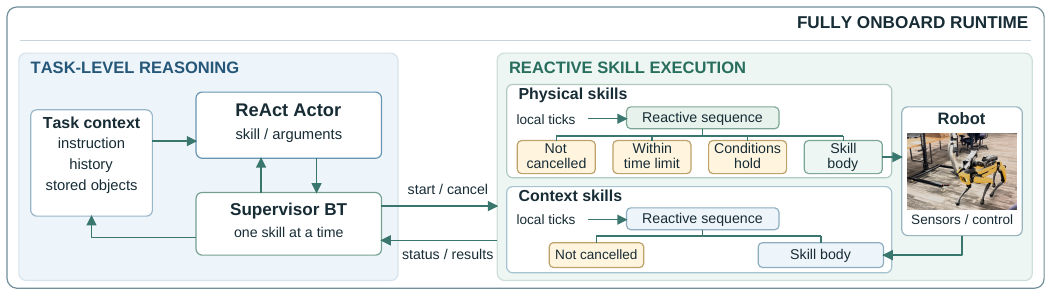}\\[1mm]
    {\small Fig.~\thefigure.\quad
    \textbf{\sysname{} separates onboard task reasoning from reactive execution.}
    The Actor selects skills, a supervisor dispatches them, and local guards enforce
    execution conditions. Observations and failure feedback guide subsequent decisions.
    }%
  }%
  \vspace{1ex}%
}

\maketitle
\AddToShipoutPictureFG*{\AtPageUpperLeft{\put(0,-0.42in){\makebox[\paperwidth]{\footnotesize
This work has been submitted to the IEEE for possible publication. Copyright may be transferred without notice, after which this version may no longer be accessible.}}}}
\thispagestyle{empty}
\pagestyle{empty}

\begin{abstract}
Deploying language-model agents onboard mobile robots requires task reasoning and responsive execution within limited compute and memory. We propose BT-Agent, an onboard architecture combining LLM-driven task decisions with predefined behavior-tree (BT) skills. These skills enforce execution conditions and return structured failure feedback. We demonstrate fully onboard operation with a 9B model on a quadruped across five task categories. Intervention tests show that local checks stop ongoing navigation without waiting for another model decision, while the agent remains active to retry or report constraints. To compare agent designs efficiently, we introduce a 161-task multi-fidelity benchmark, evaluated in text on all tasks and in Habitat on 39. Text supports 10.9× higher evaluation throughput, while simulation reveals how perception and motion failures change task outcomes. BT-Agent's largest success gains over the strongest Act/ReAct baseline occur at 4B and 9B: approximately 10 and 5 percentage points. Ablations identify structured tool calls and informative failure feedback as the most useful components.
\end{abstract}

\section{INTRODUCTION}
\label{sec:introduction}
Mobile robots increasingly receive high-level goals that require sequences of physical actions.
During search, inspection, and manipulation, they need to track partial
information and adapt when an action fails \cite{pirk2021longhorizon}. For
operation independent of remote compute, this reasoning must run onboard,
sharing finite memory and compute with perception and control. Deploying an
agent on an embedded platform therefore involves both task-level decision
making and responsive physical execution.

Language-model robot agents interleave task-level reasoning with action through
skill selection, execution feedback, and programs over robot APIs
\cite{ahn2022saycan,huang2022inner,liang2023codeaspolicies,fu2026capx}. These
methods primarily address what the agent should do next. There is, however, no
shared interface across these approaches for how selected robot skills expose
execution state, enforce conditions, and support interruption. Behavior trees
(BTs) provide a modular, robotics-native basis for such an
interface~\cite{colledanchise2018behavior}.

To this end, we propose \sysname{}, a deployment-oriented architecture that
separates task-level model reasoning from reactive robot execution through
predefined BT skills (Fig.~\ref{fig:architecture}). An onboard Actor selects
skills and arguments using the task, state, and previous observations. A
supervisor BT starts each skill and returns structured feedback, forming a
reasoning-and-acting (ReAct) loop~\cite{yao2023react}. Within each skill, guards
recheck conditions on every tick and can interrupt a running action without
another model decision. The Actor cannot modify skill implementations or
guards, so their logic can be reviewed and tested independently of the
task-level policy. This combines established BT execution mechanisms with
flexible, feedback-driven task reasoning: the model adapts its decisions,
while the BT enforces the conditions under which each skill can run.

Evaluating this architecture means checking both the agent's decisions and
how its skills perform. We need controlled comparisons across models and
harnesses, alongside trials where perception and motion can fail. We therefore
construct a multi-fidelity benchmark with 161 text tasks derived from Habitat
and Isaac scenes, sharing task goals and skill interfaces with simulator
execution. Text trials support inexpensive
comparisons of task performance, inference cost, and responses to injected
faults such as navigation timeouts and observation failures. Paired trials on
39 primary Habitat tasks examine where perception and motion change the outcome. Robosuite
experiments~\cite{zhu2020robosuite} additionally examine task-level orchestration
in manipulation. Finally, real-robot trials demonstrate fully onboard task
completion with a 9B model, while intervention tests validate the architecture's
reactive skill execution.

In summary, our contributions are:
\begin{enumerate}
    \item[i.] An agentic robot architecture, \sysname{}, for onboard deployment, combining
    model-driven task decisions with predefined BT skills, reactive execution
    checks, and structured feedback.
    \item[ii.] A multi-fidelity evaluation methodology connecting a 161-task
    text benchmark, simulation-informed fault injection, and physical execution
    through shared task goals and skill contracts.
    \item[iii.] An empirical evaluation across model sizes and execution
    backends, characterizing task success and inference cost, demonstrating
    fully onboard deployment with a 9B model, and achieving
    10.9$\times$ higher evaluation throughput in text than in Habitat at eight
    concurrent workers.
\end{enumerate}

We will release the \sysname{} implementation and benchmark code under the
Apache-2.0 license.

\section{RELATED WORK}
\textbf{Behavior trees for robot execution.}
Behavior trees (BTs) organize robot behavior into modular, reactive control
structures \cite{colledanchise2018behavior}. PA-BT combines
planning and acting by expanding trees during execution
\cite{colledanchise2019pabt}. VLM-BT generates trees with visual condition
nodes \cite{wake2025vlmbt}, while BTGenBot-2 uses a small model to generate
and repair executable trees \cite{izzo2026btgenbot2}. Ahmad et al. combine
VLM failure analysis with reactive BT planning \cite{ahmad2025unified}.
Contract-grounded synthesis validates generated trees against robot-reported
skills and permitted operators before execution \cite{salfity2026contract}. Our onboard agent selects fixed BT skills with local checks and
interruption support (Table~\ref{tab:related-comparison}).

\textbf{Agentic systems and onboard deployment.}
Language-model robot systems select skills or compose programs using task and
execution feedback \cite{ahn2022saycan,huang2022inner,liang2023codeaspolicies,fu2026capx}.
Thea combines callable robot tools with scene context and action evaluation
\cite{wang2026thea}; RoboHarness uses execution memory to coordinate
heterogeneous policies \cite{huang2026roboharness}. ROSClaw provides
model-independent ROS integration with pre-execution action validation
\cite{cardenas2026rosclaw}. These agent-planning studies do not report evaluation of their
task-level decision model under robot-borne compute limits; Thea identifies onboard inference
as future work \cite{wang2026thea}. Explicit onboard deployments are still
limited among the systems reviewed here. PRISM runs distilled planners onboard
ground and aerial robots \cite{ravichandran2025prism}, while Kwon et al.
demonstrate on-device VLM interaction on a quadruped \cite{kwon2025ondevice}.

\textbf{Multi-fidelity robot evaluation.}
BEHAVIOR-1K supports realistic physical evaluation of everyday activities
\cite{li2023behavior1k}, while Embodied Agent Interface uses symbolic
representations of BEHAVIOR and VirtualHome tasks to study embodied decision
making \cite{li2024eai}. ALFWorld connects text interaction with visually
grounded execution \cite{shridhar2021alfworld}. Building on these precedents, we align task goals and skill interfaces
across text and simulation while retaining backend-specific perception,
dynamics, and control.

\begin{table*}[t]
\centering
\caption{Features of cited configurations. Skill contracts: declared preconditions and per-call result/status semantics (partial: incomplete contract);
Guards: enforced constraints; Agentic: feedback-driven model decisions;
Recovery: post-failure correction; Onboard: robot-borne inference.
\tblYes: reported; \tblPartial: partial; \tblNo: not reported/not included; --: not applicable.
Partial guards use interface validation or model-mediated checks; PRISM varies by planner.}
\label{tab:related-comparison}
{\fontsize{8}{9.5}\selectfont
\setlength{\tabcolsep}{2pt}
\renewcommand{\arraystretch}{1.0}
\begin{tabular}{@{}l*{15}{c}@{}}
\toprule
 & \multicolumn{5}{c}{BT planning / execution} & \multicolumn{8}{c}{LLM robot agents} & \multicolumn{1}{c}{Onboard} & \\
\cmidrule(lr){2-6}\cmidrule(lr){7-14}\cmidrule(lr){15-15}
Feature & \shortstack{Wake\\et al.\\{}\cite{wake2025vlmbt}} & \shortstack{Ahmad\\et al.\\{}\cite{ahmad2025unified}} & \shortstack{BTGen-\\Bot-2 (ER)\\{}\cite{izzo2026btgenbot2}} & \shortstack{Contract-\\grounded\\BT\\{}\cite{salfity2026contract}} & \shortstack{PA-BT\\{}\cite{colledanchise2019pabt}} & \shortstack{SayCan\\{}\cite{ahn2022saycan}} & \shortstack{Inner\\Monologue\\{}\cite{huang2022inner}} & \shortstack{CaP-\\Agent0\\{}\cite{fu2026capx}} & \shortstack{ROSClaw\\{}\cite{cardenas2026rosclaw}} & \shortstack{Thea\\{}\cite{wang2026thea}} & \shortstack{Robo-\\Harness\\{}\cite{huang2026roboharness}} & \shortstack{VLAs-as-\\Tools\\{}\cite{lei2026vlastools}} & \shortstack{Cortex\\{}\cite{peng2026cortex}} & \shortstack{PRISM\\{}\cite{ravichandran2025prism}} & \shortstack{\textbf{Ours}} \\
\midrule
Skill contracts & \tblPartial & \tblYes & \tblPartial & \tblPartial & \tblYes & \tblPartial & \tblPartial & \tblPartial & \tblPartial & \tblYes & \tblPartial & \tblPartial & \tblPartial & \tblPartial & \tblYes \\
BT & \tblYes & \tblYes & \tblYes & \tblYes & \tblYes & \tblNR & \tblNR & \tblNR & \tblNR & \tblNR & \tblNR & \tblNR & \tblNR & \tblNR & \tblYes \\
Guards & \tblPartial & \tblPartial & \tblPartial & \tblPartial & \tblYes & \tblNR & \tblNR & \tblNR & \tblYes & \tblYes & \tblNR & \tblPartial & \tblPartial & \tblPartial & \tblYes \\
Agentic & \tblPartial & \tblYes & \tblYes & \tblNo & -- & \tblNo & \tblYes & \tblYes & \tblYes & \tblYes & \tblYes & \tblYes & \tblYes & \tblPartial & \tblYes \\
Recovery & \tblPartial & \tblYes & \tblYes & \tblYes & \tblYes & \tblNR & \tblYes & \tblYes & \tblNR & \tblYes & \tblYes & \tblYes & \tblYes & \tblNR & \tblYes \\
Onboard & \tblNR & \tblNR & \tblNR & \tblNR & -- & \tblNR & \tblNR & \tblNR & \tblNR & \tblNR & \tblNR & \tblNo & \tblNR & \tblYes & \tblYes \\
\bottomrule
\end{tabular}
}

\end{table*}

\section{BT-AGENT}
\label{sec:methodology}
\label{sec:method}
An LLM-driven robot should enforce execution conditions even when the model
proposes an unsuitable action. \sysname{} therefore separates task-level choice
from execution: the agent selects actions, while predefined BT skills determine
whether and how they execute (Fig.~\ref{fig:architecture}).

\subsection{Agent}
\label{sec:agent}
The LLM Actor reasons about the next step, selects a skill, and uses the
returned observation to inform its next decision. This reasoning-and-acting
loop follows ReAct~\cite{yao2023react}. Using the task, history, and stored
results, it selects skill arguments or claims completion. A supervisor BT
executes one skill at a time, returning status and results before the next
Actor call. The supervisor converts the Actor's completion claim into a terminal
submission. The Actor receives text and structured data; perception skills process
images.

We summarize history to limit context and retain skill results in a typed object
store for reuse. Failure feedback lets the Actor retry, change actions,
or report why it cannot continue. Cancelling the task interrupts its running
skill.

\subsection{BT Skills}
\label{sec:bt-skills}
We implement each robot skill as a predefined behavior tree and expose it as a
named tool. Developers can review its fixed execution logic independently of
model choices.

Consider a request to photograph a pair of glasses. The Actor queries scene
memory, navigates to a returned viewpoint, surveys the cameras, and captures
the glasses. Query and survey provide context; navigation and capture perform
task actions.

We keep context skills available so the Actor can gather information and report
constraints even when physical actions are blocked.

We check preconditions before allowing a physical skill to start. Navigation
requires a known target location. The framework validates arguments, resolves
object references, and checks required prior observations. These conditions
filter available physical skills and reject ineligible calls before execution.

Guards determine whether execution may continue: localization can fail after
navigation starts. Each physical skill's reactive sequence rechecks cancellation,
elapsed time, and skill-specific guards on every tick, a periodic tree update.
By construction, the skill body advances only if all guards pass on that tick.
A failed guard interrupts the branch and requests that the controller stop.
Checks run without further Actor calls, and the Actor cannot disable them through
skill arguments. We also check cancellation in context skills.

Failed guards or leaves supply diagnostic feedback when available, without
another model call. Physical response depends on the specified checks, state
feedback, and controller stopping behavior.

BT-Agent does not prescribe a fixed skill library. For our inspection benchmark,
we use the skills described in Section~\ref{sec:protocol}.

\section{SIMULATION FRAMEWORK}
\label{sec:protocol}
\begin{figure*}[t]
\centering
\includegraphics[trim=0bp 2.2bp 0bp 11.7bp,clip,width=\textwidth]{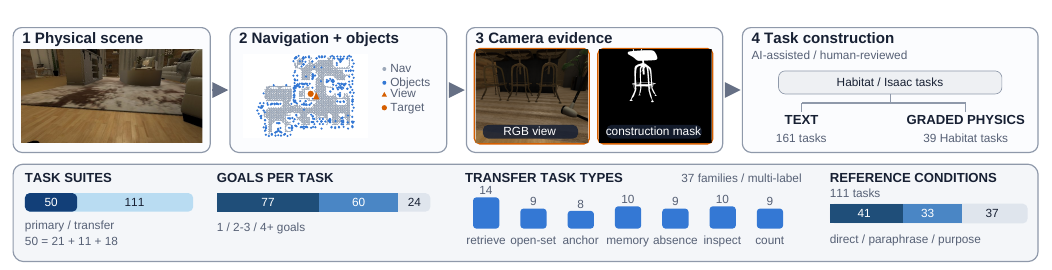}
\caption{Physics-aligned task construction. Simulator geometry and camera records produce scene memory, navigation graphs, and task definitions. Habitat and Isaac tasks receive text counterparts; graded simulator evaluation covers 39 primary Habitat tasks. Instance masks are construction evidence, not VLM input.}
\label{fig:task-distribution}
\end{figure*}

\begin{figure*}[t]
\centering
\includegraphics[width=\textwidth]{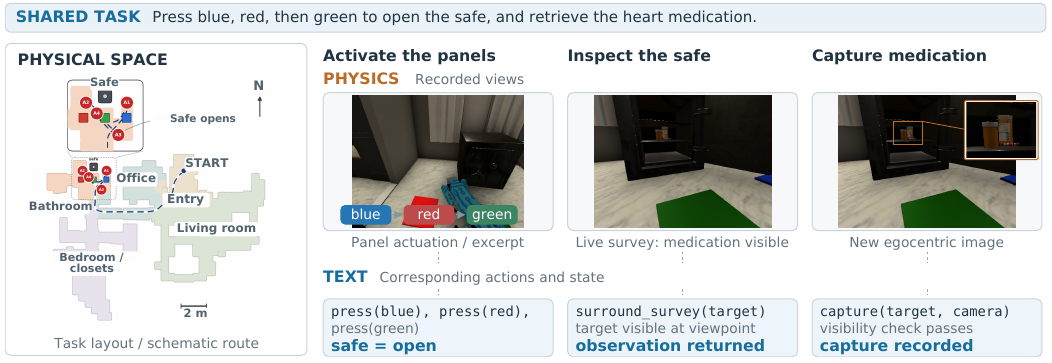}
\caption{Shared safe task across physics and text. Physical views combine
an interaction excerpt with scripted survey and capture. The map route is
schematic; the capture is displayed upright with an enlarged target.}
\label{fig:paired-task}
\end{figure*}

Physics simulation makes repeated tests of agent decisions and skill execution
costly. We extend externally supplied Habitat and Isaac environments with text
counterparts for inexpensive evaluation. Shared tasks, skill interfaces, and
explicit goal checks align the backends, while text also supports controlled
fault injection. Figure~\ref{fig:task-distribution} outlines construction;
Fig.~\ref{fig:paired-task} illustrates the shared task requirements.

\subsection{Scene Coverage}
The framework covers four scene configurations: two indoor Habitat
configurations of the same home, an outdoor Isaac scene, and an additional HSSD
home~\cite{khanna2024hssd}. An independent evaluation team supplied 32 tasks for
the first three configurations to test object search, inspection, and
interaction: 21 indoor and 11 outdoor. We add 18 indoor tasks varying reference
ambiguity, target count, interaction dependencies, and infeasibility. Together
these form the 50-task primary suite (39 Habitat-derived, 11 Isaac-derived).
This suite supports prompt and harness development. We author 111 transfer
tasks in the additional home to test the resulting configurations, with model
weights unchanged.

\subsection{Aligned Execution}
For each scene, the framework provides a scene memory with object descriptions
and locations, a navigation graph, and viewpoints from which objects can be
observed. These are derived from simulator geometry and rendered images.
Collision checks validate graph connections, and object masks record visibility.
Additional records describe changes after scripted interactions. Scene memory
provides prior knowledge about the environment. The agent's own object store
and history accumulate results and observations during the current task.

Text execution uses these records to update robot position, visibility, and
interaction state. Physics simulation executes motion and renders current camera
images. In Habitat, \texttt{surround\_survey} and capture use VLM perception,
while CMU's depth-based local planning stack~\cite{cao2022aede} executes navigation
along the supplied graph. Unlike stored text observations, VLM outputs can miss
or misidentify objects. Perception and simulated motion can therefore change
the feedback and task outcome despite shared instructions and completion
requirements. Table~\ref{tab:skills} lists the skill interface.

\begin{table}[t]
\caption{Benchmark skill semantics. Blue: context; orange: physical; green: terminal.}
\label{tab:skills}
\centering
\footnotesize
\setlength{\tabcolsep}{3pt}
\begin{tabular}{@{}>{\raggedright\arraybackslash}p{0.19\columnwidth}>{\raggedright\arraybackslash}p{0.24\columnwidth}>{\raggedright\arraybackslash}p{0.51\columnwidth}@{}}
\toprule
Skill & Inputs & Effect \\
\midrule
\ctx{query} & description, exclusions & look up objects in scene memory \\
\ctx{surround\_}\newline\ctx{survey} & target & inspect five cameras; reveal visible objects and additional scene information \\
\act{navigate} & object, viewpoint & move to a stored object viewpoint \\
\act{capture} & object, camera & record photographic evidence \\
\act{press} & object & activate a reachable interaction target \\
\act{push} & object & move an object toward its configured target \\
\ctx{report} & text & record a finding, count, or absence \\
\term{submit} & text & end the episode for grading \\
\bottomrule
\end{tabular}
\end{table}

\subsection{Task Design}
Tasks combine object search and visual documentation, inspection and reporting,
physical interaction, and reporting absent or unreachable targets. Most
object-finding requests, including those phrased as ``retrieve,'' require a new
egocentric capture during the episode.
Physical simulation saves an image from the robot's current camera view.
In text, capture checks visibility at the current viewpoint and records the
target as captured if successful, without rendering an image. Stored
scene-memory images alone do not satisfy these goals.

The primary suite includes eleven tasks requiring pressing or pushing and
three requiring a report of infeasibility. The transfer suite has 37
families with three instruction variants using direct names, paraphrases, or
purpose-based references. Targets and goals stay fixed within each family.
Ten families withhold targets from queries until a local survey reveals them.
Inspection tasks require additional observations, such as reading and reporting
the word printed on a bench cushion. Other tasks require counting or reporting
absence. Ordered interactions remain in the primary suite.
Table~\ref{tab:suite} gives examples.

AI assistance supports task proposals, object descriptions, and wording
variants. Human review checks meaning, accepted targets, and visual evidence.
Each task includes completion and partial-credit rules and a reference action
sequence. A scripted oracle executes the reference sequence in text to check the
task and its goal rules. Separate checks verify that required information can be
observed. Reference action counts inform episode budgets. Agents receive
instructions and skill observations, not the scorer's target lists or goal
predicates. Captures from physics simulation are reviewed for target identity and caption
accuracy. Bounding-box precision is not part of task scoring.

\begin{table}[t]
\centering
\caption{Example instructions and required evidence (shortened). The first four
are from the additional home; the last is from the primary suite.}
\label{tab:suite}
\footnotesize
\setlength{\tabcolsep}{3pt}
\renewcommand{\arraystretch}{1.0}
\begin{tabular}{@{}>{\raggedright\arraybackslash}p{0.60\columnwidth}>{\raggedright\arraybackslash}p{0.36\columnwidth}@{}}
\toprule
Example instruction & Required evidence \\
\midrule
Photograph something used to seat a child for a meal. & New image of an appropriate seat. \\
Find the stool near the round mirror and photograph it. & Survey reveals the target; new image. \\
Photograph the bench cushion and report the word printed on it. & New image and a report of the word. \\
Photograph every black leather-and-brass bar chair; report how many. & Three distinct captures and a report of three. \\
Find the kayak for the lake outing. & Report that the kayak cannot be found. \\
\bottomrule
\end{tabular}
\end{table}

\subsection{Fault Injection}
Nominal task success alone does not show how an agent handles failure. The text
backend applies repeatable faults to eligible skill calls
(Table~\ref{tab:injected-faults}), changing execution state, returned observations,
or both. Some explicitly report failure; others report success despite incomplete
execution, testing whether agents check outcomes beyond the returned status.
Cases are informed by simulation failures and supplemented with targeted stress
cases. Matched nominal and fault-injected conditions retain the same tasks,
completion requirements, and agent settings. Section~\ref{sec:experiments}
specifies the evaluated configurations and budgets.

\begin{table}[t]
\centering
\caption{Mechanisms in the text fault schedule, grouped by effect. Faults fire
only on eligible skill calls.}
\label{tab:injected-faults}
\footnotesize
\setlength{\tabcolsep}{3pt}
\renewcommand{\arraystretch}{1.0}
\begin{tabular}{@{}>{\raggedright\arraybackslash}p{0.27\columnwidth}>{\raggedright\arraybackslash}p{0.69\columnwidth}@{}}
\toprule
Fault group & Injected behavior \\
\midrule
Localization & Stale or missing pose; unreported position jump. \\
Navigation progress & No movement, timeout, or arrival short of the target. \\
Route obstruction & Reported blockage, temporary wedge, disabled edge, or unavailable route. \\
Viewpoint & Incorrect viewpoint despite reported arrival. \\
Interaction & Press stops short without activating the target. \\
Perception & Capture analysis fails to produce evidence. \\
\bottomrule
\end{tabular}
\end{table}

\section{EXPERIMENTS}
\label{sec:experiments}
We organize the evaluation around three questions:
(\textcolor{RoleContext}{\texttt{Q1}}) Which model and harness choices support
task execution under onboard compute constraints?
(\textcolor{RoleContext}{\texttt{Q2}}) How does aligned text and simulator
evaluation support development toward deployment?
(\textcolor{RoleContext}{\texttt{Q3}}) Can BT-Agent execute tasks fully onboard
while enforcing local skill conditions and handling interruptions?

\subsection{Evaluation Setup}
We compare BT-Agent with three baselines: Act, ReAct, and adapted CaP-X.
Act selects a textual skill command from the task and history.
ReAct~\cite{yao2023react} adds reasoning before the command. Adapted
CaP-X~\cite{fu2026capx} generates and revises Python programs, retaining
persistent program state and using benchmark-specific skills, prompts, and budgets.

Methods share instructions, skills, and execution checks within each backend,
with method-specific prompts, interfaces, and history handling. These comparisons
assess task-level orchestration over a common skill layer. The onboard trials
separately test the architecture's end-to-end operation and reactive skill execution.

\subsection{Model Size and Harness Design (\textcolor{RoleContext}{\texttt{Q1}})}
We compare agent configurations on the 50-task primary and 111-task transfer
suites with shared skills and execution checks, using three paired seeds per
task in both main comparisons and nominal ablations. Models are Qwen3.5 NVFP4 checkpoints at 4B, 9B, 27B, and
122B-A10B.\footnote{Text sweeps use vLLM on B300 GPUs with greedy decoding.}

\label{sec:budget-regime}
Episodes allow three times the reference operation count and 1,800\,s.
Act, ReAct, and BT-Agent receive 1,024 output tokens per action-selection
call; adapted CaP-X receives 20,480 per program. BT-Agent's separate
history-summarization calls are capped at 512 output tokens.

\label{sec:metrics}
We report success and median/P99 output tokens per episode, including failures and
context exhaustion. Token counts use returned model usage, excluding
unavailable retry usage. Median and P99 capture typical and unusually
expensive episodes.

\begin{table}[t]
\centering
\caption{Nominal/fault success (SR, \%) and nominal output (k tokens), including failures. SR161 is task-weighted; P99: 99th percentile. Bold/underline: best/second per block, model, and metric. W: watchdog; —: unavailable. Bottom: matched nominal ablations with a fresh control. Both blocks use three seeds.}
\label{tab:q3-nominal}
\fontsize{8.5}{9.5}\selectfont
\renewcommand{\arraystretch}{0.95}
\setlength{\tabcolsep}{1.9pt}
\sbox0{%
\begin{tabular}{@{}llrr>{\columncolor{TableAll}}rr>{\columncolor{TableAll}}r>{\columncolor{TableAll}}r@{}}
\toprule
 & & \multicolumn{3}{c}{Nominal SR$\uparrow$} & \multicolumn{2}{c}{Output (k)$\downarrow$} & Fault SR$\uparrow$ \\
\cmidrule(lr){3-5}\cmidrule(lr){6-7}\cmidrule(l){8-8}
Model & Method & 50 & 111 & 161 & Median & P99 & 161 \\
\midrule
4B & Act & 45.3 & 31.2 & 35.6 & \textbf{0.20} & \textbf{2.1} & 28.6 \\
 & ReAct & 40.0 & \underline{42.0} & 41.4 & 1.20 & 36.1 & \underline{34.8} \\
 & CaP-X (adapted) & 2.7 & 1.2 & 1.7 & 21.49 & 61.4 & 0.6 \\
 & BT-Agent & \textbf{70.7} & \textbf{42.9} & \textbf{51.6} & 1.09 & 20.9 & \textbf{42.2} \\
 & BT-Agent+W & \underline{68.0} & \textbf{42.9} & \underline{50.7} & \underline{1.07} & \underline{13.4} & — \\
\midrule
9B & Act & 55.3 & 36.9 & 42.7 & \textbf{0.24} & 42.1 & 33.7 \\
 & ReAct & 60.7 & 44.7 & 49.7 & \underline{0.85} & 29.3 & \underline{45.8} \\
 & CaP-X (adapted) & 32.7 & 11.1 & 17.8 & 12.63 & 48.1 & 9.9 \\
 & BT-Agent & \underline{66.0} & \textbf{49.5} & \underline{54.7} & 1.06 & \underline{28.4} & \textbf{52.4} \\
 & BT-Agent+W & \textbf{72.7} & \underline{48.0} & \textbf{55.7} & 1.10 & \textbf{16.6} & — \\
\midrule
27B & Act & \textbf{79.3} & 55.0 & \underline{62.5} & \textbf{0.14} & \textbf{1.6} & 54.2 \\
 & ReAct & \underline{78.0} & 56.2 & \textbf{62.9} & \underline{0.65} & \underline{5.5} & \underline{54.9} \\
 & CaP-X (adapted) & 45.3 & 32.1 & 36.2 & 7.08 & 41.0 & 21.9 \\
 & BT-Agent & 74.0 & \textbf{58.0} & \textbf{62.9} & 1.05 & 10.1 & \textbf{56.7} \\
 & BT-Agent+W & 74.7 & \underline{56.8} & 62.3 & 1.01 & 9.0 & — \\
\midrule
\shortstack[l]{122B-\\A10B} & Act & 73.3 & 57.1 & 62.1 & \textbf{0.16} & 22.4 & 54.2 \\
 & ReAct & \underline{79.3} & 59.5 & 65.6 & \underline{0.83} & \underline{19.3} & \underline{55.3} \\
 & CaP-X (adapted) & 60.0 & 35.4 & 43.1 & 6.13 & 41.0 & 31.7 \\
 & BT-Agent & 78.0 & \textbf{63.1} & \underline{67.7} & 1.06 & 22.5 & \textbf{59.2} \\
 & BT-Agent+W & \textbf{80.0} & \underline{62.5} & \textbf{67.9} & 1.04 & \textbf{16.6} & — \\
\bottomrule
\end{tabular}%
}
\typeout{Q3TAIL-WIDTH=\the\wd0; COLUMN-WIDTH=\the\columnwidth}
\ifdim\wd0>\columnwidth\PackageError{Q3-table}{Table exceeds column width}{Review layout}\fi
\dimen0=\wd0
\usebox0

\vspace{2pt}
\sbox0{%
\begin{tabular*}{\dimen0}{@{\extracolsep{\fill}}lrrrr@{}}
\toprule
\multicolumn{5}{@{}l}{Matched BT-Agent ablations: nominal SR161$\uparrow$} \\
Variant & 4B & 9B & 27B & 122B \\
\midrule
BT-Agent (control) & \textbf{50.5} & \textbf{58.0} & \underline{62.5} & \underline{66.7} \\
$-$ capture-specific guidance & \underline{49.5} & \underline{55.5} & \textbf{63.1} & \textbf{67.9} \\
Text-action interface & 37.5 & 44.3 & 54.7 & 59.6 \\
Opaque failure feedback & 46.2 & 52.6 & 60.5 & 60.7 \\
\bottomrule
\end{tabular*}%
}
\usebox0
\end{table}

\subsubsection{Nominal task performance}
BT-Agent matches or exceeds the strongest Act/ReAct baseline at every size
on the combined suite, with the largest gains at 4B and 9B: 10.1 and 5.0
percentage points (Table~\ref{tab:q3-nominal}). Transfer gains are positive
across sizes and largest at 9B, at 4.8 points over ReAct, while the larger
4B gain comes mainly from the primary suite.

In 4B text traces of the safe task, ReAct visits the panels
but submits without pressing them, while BT-Agent completes the sequence
and photographs the medication. In an inscription task (Table~\ref{tab:suite}),
both identify the printed word, but only BT-Agent records it using the report skill.
These cases suggest making required actions and procedures explicit in
harness design, especially for smaller models.

\subsubsection{Component ablations}
Matched ablations use a fresh control with the controller, skills, checks,
and tasks fixed. Replacing structured tool calls with textual commands in
action selection and history lowers success at every size, most at 4B and
9B, where median output rises $1.9\times$ and $1.7\times$ on nominal tasks.

Reducing failure feedback to a skill name and status lowers nominal success by
2--6 points across sizes. BT execution exposes failed guards or leaves,
providing diagnostics as a by-product without another model call.
Removing capture-specific guidance has little effect. These ablations show
that the action interface and informative failure feedback contribute to
task success, with smaller models particularly sensitive to harness design.

\subsubsection{Inference cost and fault robustness}
Act has the lowest median output, consistent with direct action selection
without reasoning. At 4B, BT-Agent reduces ReAct's P99 output by 42\% across
161 tasks. Simpler interfaces become more competitive at larger sizes:
27B ReAct matches BT-Agent's nominal success with lower median and P99 output.

The optional response watchdog stops episodes after repeated unusable outputs,
repeated equivalent queries or failed-skill outcomes, or reaching 16,384 output tokens
or 128 model calls. It lowers P99 at every size with minor impact on success.
Token checks follow complete responses, allowing small overshoots.

Under the faults in Table~\ref{tab:injected-faults}, BT-Agent has the highest
observed success at every size. Matched no-fault controls keep each agent's
skills and settings fixed. At 27B, the nominal tie with ReAct becomes a
BT-Agent lead under faults. Nominal performance can therefore conceal
differences in failure handling, motivating fault testing for deployment.

A paired 4B breakfast trace illustrates the difference. Both methods succeed
without faults and receive identical observations up to an injected route
blockage. BT-Agent changes viewpoint and completes, whereas ReAct retries the
blocked approach and later repeats unsuccessful captures until exhausting its
budget.

\subsubsection{Manipulation transfer}
We use Robosuite~\cite{zhu2020robosuite} to compare program generation with
stepwise skill selection in manipulation. Here, skills use blocking API calls
rather than BT execution. We evaluate six reference tasks from
CaP-X~\cite{fu2026capx} and eight sorting, stacking, and rearrangement
extensions, with 20 repetitions per task at 4B, 9B, and 27B.\footnote{We omit the reference
cube-restacking task, reported at this tier for only one model. Precise insertion
is also excluded: eight of nine code-policy successes exceed the 1,000-step
simulator limit, which is checked only between program generations.}
CaP-X generates programs over its raw API or our shared skills. Act, ReAct,
and BT-Agent invoke the shared skills one action at a time. Raw-API runs use
ground-truth poses instead of perception, with either a single program or
multi-turn revision. Stepwise agents retain our reference-based action
budget, while code policies have no action-count cap.

BT-Agent leads combined success at all three sizes through its advantage
on the extensions (Table~\ref{tab:harness-comparison}), while shared-skill
CaP-X leads the reference suite at 9B and 27B. On the six reference tasks,
our 27B raw-API single-turn policy reaches 44.2\%, within the original paper's
31.2--65.7\% range across twelve models (20B: 51\%; human programs: 94\%).

Individual tasks reveal a trade-off between program generation and stepwise
decisions. Wiping follows a regular 101-call raster that a program expresses
as a loop. At 27B, shared-skill multi-turn CaP-X completes 14/20 trials,
while BT-Agent completes none within its time budget. Three-cube restacking
requires responding to the observed stack order. BT-Agent completes 15/20
trials versus 4/20 for shared-skill CaP-X. These cases suggest using programs
for long, regular sequences and stepwise decisions when observations determine
the next action.
\begin{table}[t]
\centering
\caption{Robosuite SR (\%), 20 repetitions/task. O6: six CaP-X reference tasks; A8: eight extensions; All: fourteen tasks. CaP-X uses single- or multi-turn programs over its privileged API or shared skills, without an action-count cap. Bold/underline: best/second per column, including ties.}
\label{tab:harness-comparison}
\footnotesize
\setlength{\tabcolsep}{2.0pt}
\begin{tabular}{@{}lrr>{\columncolor{TableAll}}rrr>{\columncolor{TableAll}}rrr>{\columncolor{TableAll}}r@{}}
\toprule
 & \multicolumn{3}{c}{4B} & \multicolumn{3}{c}{9B} & \multicolumn{3}{c}{27B} \\
\cmidrule(lr){2-4}\cmidrule(lr){5-7}\cmidrule(lr){8-10}
 & O6 & A8 & All & O6 & A8 & All & O6 & A8 & All \\
\midrule
\multicolumn{3}{@{}l}{\emph{Raw privileged API}} & & & & & & & \\
CaP-X single & 17.5 & 8.1 & 12.1 & 26.7 & 2.5 & 12.9 & 44.2 & 47.5 & 46.1 \\
CaP-X multi & 7.5 & 11.9 & 10.0 & 16.7 & 4.4 & 9.6 & \underline{50.8} & 55.6 & 53.6 \\
\multicolumn{3}{@{}l}{\emph{Shared skill library}} & & & & & & & \\
CaP-X single & 16.7 & 4.4 & 9.6 & 32.5 & 36.9 & 35.0 & 42.5 & 45.6 & 44.3 \\
CaP-X multi & 9.2 & 25.0 & 18.2 & \textbf{37.5} & 46.2 & 42.5 & \textbf{54.2} & 62.5 & 58.9 \\
ReAct & \underline{32.5} & 39.4 & 36.4 & \underline{35.0} & 62.5 & 50.7 & 35.8 & 76.9 & 59.3 \\
Act & \textbf{35.0} & \underline{60.0} & \underline{49.3} & 34.2 & \underline{70.0} & \underline{54.6} & 36.7 & \underline{78.8} & \underline{60.7} \\
\sysname{} & \textbf{35.0} & \textbf{75.6} & \textbf{58.2} & 34.2 & \textbf{74.4} & \textbf{57.1} & 36.7 & \textbf{81.9} & \textbf{62.5} \\
\bottomrule
\end{tabular}
\end{table}

\begin{figure}[t]
\centering
\includegraphics[width=\columnwidth]{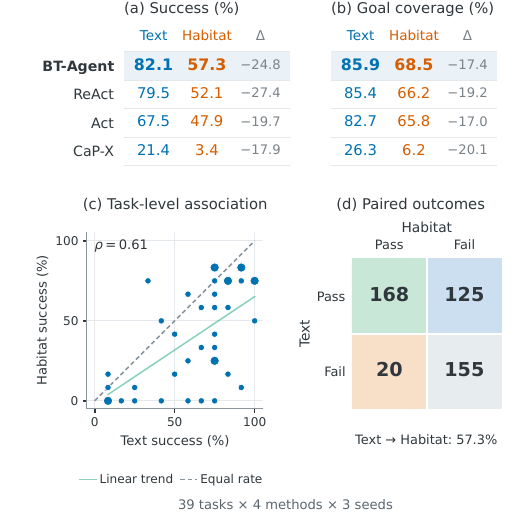}
\caption{Text--Habitat comparison: 39 tasks, 117 episodes/method. (a,b) Bold: highest observed; gray $\Delta$: Habitat minus text (percentage points). (c) Descriptive association of task means over four methods and three seeds in two configurations of one home; marker size: coincident tasks; $\rho$: Spearman correlation. (d) Pooled paired outcomes: 168/293 text successes also succeed in Habitat.}
\label{fig:q2-paired-9b}
\end{figure}

\begin{figure*}[!t]
\centering
\includegraphics[width=\textwidth]{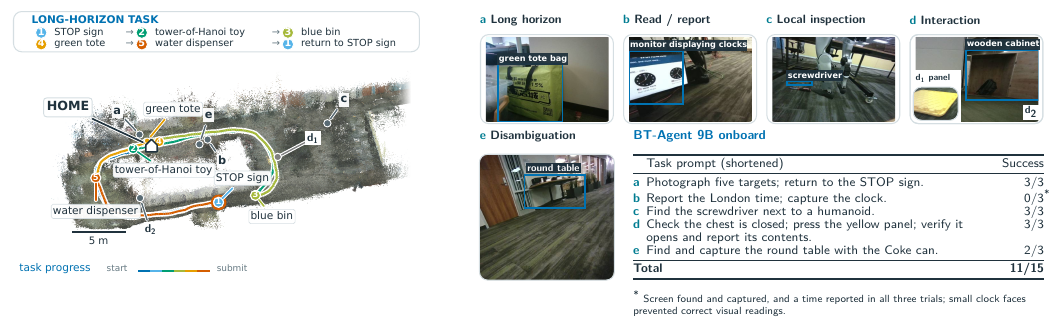}
\caption{Onboard deployment across five tasks. Left: task locations with the long-horizon trajectory and capture points. HOME: start/return pose; d\textsubscript{1}/d\textsubscript{2}: yellow panel/chest. Right: recorded observations, shortened prompts, and operator-adjudicated successes over three trials/task.}
\label{fig:q1-deployment}
\end{figure*}

\begin{figure}[t]
\centering
\includegraphics[trim=0bp 10.5bp 0bp 1.6bp,clip,width=\columnwidth]{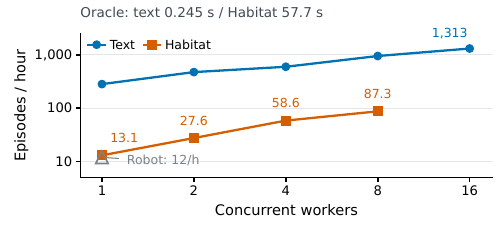}
\caption{Oracle mean time and agent throughput on 16 Habitat tasks; logarithmic axes. Throughput excludes timeouts and ends at the last non-timeout completion. Habitat uses one RTX 5090, with an asynchronously batched 9B VLM on one RTX PRO 6000. Triangle: serial real-robot execution, excluding setup, charging, and debugging.}
\label{fig:q2-cost-throughput}
\end{figure}

\begin{figure}[t]
\centering
\includegraphics[width=\columnwidth]{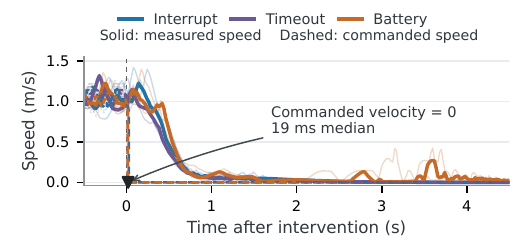}
\caption{Reactive stopping across nine interventions. Commanded velocity reaches zero a median 19\,ms after local acceptance; measured speed reflects subsequent locomotion control and physical settling. Battery trials also include sitting down.}
\label{fig:q1-bt-probes}
\end{figure}

\subsection{From Text to Simulator Evaluation (\textcolor{RoleContext}{\texttt{Q2}})}
\subsubsection{Paired text--Habitat outcomes}
We compare text and Habitat on 39 primary tasks in two configurations of
the same home, using four agent configurations with a 9B model and three paired
seeds per task. Isaac-derived tasks and the transfer suite are evaluated quantitatively
in text only. The backends (Section~\ref{sec:protocol}) share instructions,
scene priors, and a 900-second budget, but agents follow their own observations.
Human review checks visual evidence; recorded state and answer checks grade
interactions and reports. Success requires all goals and a valid submission
within budget, including turn-limit submissions. Goal coverage measures partial
completion independently of submission.

The observed method ordering agrees across backends
(Fig.~\ref{fig:q2-paired-9b}). Tasks with higher success in text also tend to
have higher success in Habitat (Spearman $\rho=0.61$ across 39 task means).
Of 188 successful Habitat runs, 168 succeed in text, but only 57.3\% of text
successes succeed in Habitat. Symbolic success therefore does not establish
completion with sensing and motion.

Paired traces illustrate agreement and divergence. Both glasses runs follow
query--navigate--survey--capture but select different cameras. On wallet/backpack,
survey and capture disagree about wallet visibility in the same image, causing
repeated failed captures. On an absent-object task, one text run queries until
timeout, while its Habitat counterpart reports the unsuccessful lookup and submits. Differences can
therefore arise in decisions and stopping behavior as well as perception and motion.

\subsubsection{Evaluation cost}
On 16 tasks, oracles without model inference average 0.245 seconds per episode
in text and 57.7 seconds in Habitat with ground-truth perception.
With a shared 9B model hosted on an RTX~PRO~6000, eight workers achieve
950 episodes/hour in text versus 87.3 in Habitat on one RTX~5090: a
10.9$\times$ difference excluding timeout attempts (Fig.~\ref{fig:q2-cost-throughput}).
Agent throughput includes inference, startup and cleanup, and Habitat's visual
skills and motion.

Text supports rapid orchestration development, while paired simulator trials
reveal skill failures and changed decisions before onboard deployment.

\subsection{Onboard Operation and Reactive Skill Execution (\textcolor{RoleContext}{\texttt{Q3}})}
\label{sec:onboard-execution}
Finally, we deploy the full system onboard a mobile robot to evaluate task
execution and local skill supervision.

\subsubsection{Onboard task execution}
We deploy BT-Agent on a Boston Dynamics Spot quadruped with NVIDIA Jetson Thor,
using a 9B model for the Actor and visual skills. The agent queries a prebuilt
FARM scene graph~\cite{he2026farm} containing object descriptions, positions,
and recorded views from a prior RGB-D mapping run, with gross labeling errors
manually reviewed and corrected. Retrieval uses onboard Qwen3-Embedding-0.6B
and the same 9B model for candidate verification. A pre-generated waypoint
graph supports global planning, while adapted CMU terrain-analysis,
local-planning, and path-following modules~\cite{cao2022aede} execute motion.
All runtime inference, retrieval, BT execution, planning, and control run
onboard, while motion requires operator approval.

Across five tasks with three trials each, BT-Agent completes 11 of 15 trials
(Fig.~\ref{fig:q1-deployment}). It coordinates
multi-stop navigation and capture, uses a mapped landmark to find an unmapped
screwdriver, and checks and operates the chest through the floor panel.
In the reporting task, it finds and captures the display but reports an
incorrect time after misreading the small clock faces. Together, these trials
demonstrate end-to-end deployment of the full architecture, with reasoning,
perception, and control onboard.

\subsubsection{Local guards and interruption}
Each guard and intervention condition is tested three times, with BT traces
identifying the active skill and leaf. After a guard rejects pressing without
a survey, the Actor surveys its surroundings and retries. At a simulated
9\% battery reserve, navigation is rejected and the Actor reports the constraint
and abandons the task.

Operator interruptions and injected timeouts stop navigation while keeping the
agent active. The Actor receives timeouts as ordinary skill failures and retries
navigation after either intervention. All timeout trials complete.
A simulated 5\% reserve during navigation triggers a persistent motion hold
and the built-in sit-down command. The Actor reports the constraint without
retrying motion.

Across nine interventions during motion, commanded velocity reaches zero a
median 19\,ms after local acceptance (Fig.~\ref{fig:q1-bt-probes}), while the
next model-selected action follows after 3.2--3.9\,s in interruption and timeout
trials. Local BT checks interrupt the skill before the Actor decides how
to proceed.

\section{CONCLUSION}
\label{sec:discussion}
Onboard trials demonstrate \sysname{}'s combination of task reasoning and
reactive execution: the agent coordinates multi-step tasks, while local BT
checks enforce execution conditions and interrupt motion without waiting for
another model decision.
Smaller models benefit most from harness support, while useful components
depend on the task and inference cost. Text enables inexpensive development,
and simulator trials expose skill and decision failures.

\subsection{Limitations and Future Work}
The Actor can sequence and parameterize available skills, but their
implementations and guards stay fixed at runtime. This supports review and
testing before deployment, but new capabilities require skill authoring and
platform-specific integration. Reliably conveying task-relevant evidence
remains an open challenge when observations are incomplete or ambiguous.

Task-level reasoning could be extended to run alongside BT supervision of
ongoing skills. As edge models improve, they could also reason explicitly over
tree structure and failed leaves to diagnose failures and select more effective
recovery actions. Evaluating frontier models within the same architecture would
help distinguish model limitations from those of the skill interface. Finally,
the executable text environment could support reinforcement learning of
task-level policies, followed by evaluation in simulation and on the robot.
\section*{Acknowledgments}
This work was supported in part by the DARPA TIAMAT program.

OpenAI Codex and Anthropic Claude assisted with brainstorming, prose refinement throughout the manuscript, and research and plotting code
(Sections~\ref{sec:method}--\ref{sec:experiments} and figures). 
\FloatBarrier
\bibliographystyle{IEEEtran}
\bibliography{references}
\end{document}